\def\year{2021}\relax
\documentclass[letterpaper]{article} 
\usepackage{aaai21}  
\usepackage{times}  
\usepackage{helvet} 
\usepackage{courier}  
\usepackage[hyphens]{url}  
\usepackage{graphicx} 
\usepackage{natbib}  
\usepackage{caption} 
\usepackage{amsmath,amssymb,amsfonts,bm}
\usepackage{algorithmic}
\usepackage{xcolor}
\usepackage[linesnumbered,ruled,vlined]{algorithm2e}
\begin{document}
%
\SetKwData{OPEN}{\texttt{OPEN}}
\SetKwData{FOCAL}{\texttt{FOCAL}}
\SetKwData{SUBOPEN}{\texttt{SUBOPEN}}
\SetKwData{NONFOCAL}{\texttt{NONFOCAL}}
\SetKwData{CLOSE}{\texttt{CLOSE}}
\newcommand{\JB}[1]{\textbf{\color{blue} JB: #1}}
\newcommand{\PA}[1]{\textbf{\color{red} PA: #1}}
\newcommand{\MG}[1]{\textbf{\color{green} MG: #1}}
\newcommand{\Paragraph}[1]{\smallskip\noindent\textbf{#1}~~}
\title{Exploiting Learned Policies in Focal Search}
\author{
    Pablo Araneda\textsuperscript{\rm 1} ~~~ Matias Greco\textsuperscript{\rm 1} ~~~ Jorge A.\ Baier\textsuperscript{\rm 1,2}
    \\
}
\affiliations{
    \textsuperscript{\rm 1} Pontificia Universidad Católica de Chile, Chile\\
    \textsuperscript{\rm 2} Instituto Milenio Fundamentos de los Datos, Chile\\
    \{pharaneda, mogreco\}@uc.cl, jabaier@ing.puc.cl

}

\maketitle
\begin{abstract}
Recent machine-learning approaches to deterministic search and domain-independent planning employ policy learning to speed up search. Unfortunately, when attempting to solve a search problem by successively applying a policy, no guarantees can be given on solution quality. The problem of how to effectively use a learned policy within a bounded-suboptimal search algorithm remains largely as an open question. In this paper, we propose various ways in which such policies can be integrated into Focal Search, assuming that the policy is a neural network classifier. Furthermore, we  provide mathematical foundations for some of the resulting algorithms. To evaluate the resulting algorithms over a  number of policies with varying accuracy, we use synthetic policies which can be generated for a target accuracy for problems where the search space can be held in memory. We evaluate our focal search variants over three benchmark domains using our synthetic approach, and on the 15-puzzle using a neural network learned using 1.5 million examples. We observe that \emph{Discrepancy Focal Search}, which we show expands the node which maximizes an approximation of the probability that its corresponding path is a prefix of an optimal path, obtains, in general, the best results in terms of runtime and solution quality. 
\end{abstract}

\section{Introduction}
In the past few years, machine learning (ML) approaches have been proposed to enhance the performance of AI search and domain-independent planning. A number of these approaches could be classified as learning a heuristic estimate \cite[e.g.,][]{YoonFG06,ArfaeeZH11,ThayerDR11,FerberH020}, a policy  \cite[e.g.,][]{GroshevGTSA18,MunozFHB18,ToyerTTX20}, or a combination of both  \cite[e.g.,][]{McAleerASB19}.

Heuristic search, which was designed to exploit cost-to-go heuristics, can accommodate learned heuristics naturally. However, it is not immediately obvious how policies can be exploited in heuristic search. This is especially relevant when we seek for bounded-suboptimal solutions, rather than satisficing solutions. As an illustration, unrolling a policy may produce very poor quality solutions even if the policy is very accurate. For example a 95\%-accurate policy for the 15-puzzle would produce a 50-action, optimal solution only with a 0.077 probability. 

In this paper we investigate various ways in which may integrate learned policies within Focal Search (FS) \cite{PearlK82}, a bounded-suboptimal algorithm that allows using heuristic functions which may not be cost-to-go estimates. We assume that policies are neural network classifiers which, in their output layer, use a softmax activation. 

We propose two families of policy-based heuristics for FS: score-based, and discrepancy-based. Both types of heuristics have been used previously. Score-based heuristics have been used along with beam search for language generation \cite{wu2016googles} 
and discrepancy-based heuristics have been used in policy-enhanced planning \cite{YoonFG07b}. But none of these approaches nor others that combine search with policies \cite[e.g.,][]{ShenTTTX19, Agostinelliabs-2102-04518,DBLP:journals/corr/abs-2103-11505,DBLP:conf/nips/OrseauLLW18} provide suboptimality guarantees. Rather than devising new heuristics, the main contribution we make when integrating these heuristics into FS, is to provide a mathematical interpretation of the resulting algorithms. For example, for discrepancy-based FS, we establish a relation between discrepancy and maximization of the probability of expanding a prefix of an optimal path.

An objective of our empirical evaluation was to understand how FS would perform as the accuracy of the policy varied. Since training policies for a given target heuristic may be very time-consuming, we propose an approach to create synthetic policies, which only works for small search problems, but it allows to carry out fine-grained studies, focused on search rather than on learning. Specifically, we use this approach with 3 domains: the 8-puzzle, blocksworld, and pancake sorting. To see whether our not our results extend to larger domains, we evaluated our FS heuristics using an 87.5\%-accurate policy for the 15-puzzle, which we obtained by training on 1.5 million examples.

Our results
show that discrepancy-based FS yields the best results in terms of runtime and expansions, while score-based approaches find better-quality solutions. FS, used with policies, outperforms the bounded-suboptimal algorithm Weighted A* (WA*) \cite{Pohl70}, both in terms of expansions and in terms of solution quality when the accuracy of the policy is greater than 80\%, though in blocksworld we observed that WA* was outperformed even when using policies whose accuracy is as low as 70\%. An important conclusion is that learned policies can improve the efficiency of bounded-suboptimal search, and that its effectiveness depends on the accuracy of the (admissible) heuristic at hand.

\section{Background}
\subsection{Search Tasks and Policies}
A search graph is a tuple $G=(S,E,A,\lambda,c)$, where $S$ is a set of states $E\subseteq S\times S$, is a finite set of edges, $c:E\rightarrow \mathbb{R}^{\geq 0}$ is a cost function associating a non-negative cost with every edge, $A$ is a set of \emph{actions} and $\lambda:E\rightarrow A$ is a labeling function that associates each arc of the graph with an action. Function $\lambda$ is such that all edges that emerge from a given state are labeled with a different action; that is, for every $s\in S$, if $\lambda(s,s') = \lambda(s,s'')$, then $s'=s''$. We say that action $a$ is \emph{applicable} in state $s$ iff $\lambda(s,s')=a$, for some $(s,s')\in E$; otherwise, it is \emph{inapplicable}. If $a$ is applicable in $s$, we define $Succ(a,s)$ as the state $s'$ such that $\lambda(s,s')=a$.  Moreover, $Succ(s)$ denotes the set $\{s'\mid (s,s')\in E\}$.

A path over a search graph $G$ is a sequence of states $s_1,s_2,\ldots, s_n$, such that $(s_i,s_{i+1})\in S$, for every $i\in\{1,\ldots,n-1\}$. A path $\sigma$ is a path from $s$ to $t$ if and only if sequence $\sigma$  starts with $s$ and ends with $t$. The cost of a path $\sigma=s_1,s_2,\ldots, s_n$ is $c(\sigma)=\sum_{i=1}^{n-1} c(s_i,s_{i+1})$. A path $\sigma$ from $s$ to $t$ is \emph{optimal} if no other path from $s$ to $t$ has a cost less than $c(\sigma)$. For any state $s\in S$, we denote the cost of an optimal path from $s$ to $s_{goal}$ by $h^*(s)$.

A \emph{search task} is a tuple $P=(G,s_{start},s_{goal})$, where $G=(S,E,A,\lambda,c)$ is a search graph, $s_{start} \in S$ is the \emph{start state}, and $s_{goal}\in S$ is the \emph{goal state}. A  solution (resp.\ optimal solution) to a search task is a path (resp.\ optimal path) connecting $n_{start}$ and $n_{goal}$.  A solution $\sigma$ is \emph{$w$-suboptimal}, where $w\geq 1$, when its cost does not exceed the cost of an optimal solution multiplied by $w$.

A \emph{stochastic policy} is a function $\pi:A,S\rightarrow [0,1]$ that maps each state-action pair to a probability, and therefore $\pi(a,s)$ is such that $\sum_{a\in A}\pi(a,s)=1$, for every state $s$. Stochastic policy $\pi$ is \emph{well-defined} if and only if $\pi(a,s)=0$ for every action $a$ that is inapplicable in $s$. By \emph{unrolling} a stochastic policy $k$ times from a given state $s$ we mean repeating the following two steps $k$ times: (1) choose an action $a\in A$ using the probability distribution $\pi(\cdot,s)$ (2) assign $s$ as $Succ(a,s)$, failing if $Succ(a,s)$ is undefined.

wA \emph{deterministic policy} can always be defined from a stochastic policy $\pi$ by returning, in each state $s$, an action with maximum probability. If $\pi$ is a stochastic policy we define the deterministic policy $\delta_\pi(s)={\rm arg}\max_{a\in A} \pi(a,s)$, and we omit the subscipt when $\pi$ is clear from the context.

Let $opt(s)$ denote the set of actions that may start an optimal path from $s$ to $s_{goal}$, that is, the set that contains the actions that are such that $s,Succ(a,s)$ is the prefix of an optimal path from $s$ to $s_{goal}$. The \emph{accuracy} of a stochastic policy $\pi$ is defined as:
\begin{equation}
    acc_\pi = \frac{1}{|S|}\sum_{s\in S} [\delta_\pi(s) \in opt(s)], \label{eq:accuracy}
\end{equation}
where expression $[B]$, if $B$ is a boolean condition, evaluates to 1 if $B$ is true, and is 0 otherwise. In words, $acc_\pi$ corresponds to the percentage of states in which an action that leads to an optimal path would be chosen by the deterministic policy associated with $\pi$. Note that it is not feasible in general to compute the accuracy as defined by \eqref{eq:accuracy}, thus we use estimates of the accuracy using a sample of states in $S$.


Given a search task, a \emph{heuristic function} $h:V\rightarrow \mathbb{R}^{\geq 0}$ is such that $h(s)$ estimates the cost of a path connecting $s$ with $s_{goal}$. Heuristic function $h$ is \emph{admissible} if and only if $h(s)\leq h^*(s)$, for every $s\in S$. 
 
 We assume the reader is familiar with the Weighted A* (WA*)~\cite{Pohl70}, a generalization of the A*~\cite{HartNR68}. In particular, with the fact that WA* maintains a search frontier, called \OPEN, in which nodes are ordered using a function $f(s)=g(s)+wh(s)$, where $g(s)$ is the cost of a path from the start state to $s$, $h$ is a heuristic function, and $w$ is a real parameter not smaller than 1. WA* is guaranteed to return $w$-suboptimal solutions when $h$ is admissible.

\subsection{Focal Search}
Focal Search (FS) \cite{PearlK82} is a well-known bounded-suboptimal search algorithm. In addition to an admissible heuristic $h$, it is capable of  exploiting additional information for guiding search. It uses two priority queues: (1) $\OPEN$, which is the search frontier sorted in ascending order by $f=g+h$, where $g$ and $h$ are defined like in WA*, and (2) $\FOCAL$, which is sorted in ascending order by $h_{\FOCAL}$, a function which may or may not be a cost-to-go estimate and which should help guide search.

Like WA*, FS receives a parameter $w$ which controls the suboptimality of the returned solution. Its \FOCAL list contains every state $s$ in \OPEN such that $f(s)\leq wf_{min}$, where $f_{min}$ is the minimum $f$-value of a node in \OPEN. 

The pseudocode of FS is presented in Algorithm~\ref{algo:fs}. In each iteration, a node $s$ is extracted from \FOCAL; $s$ is also removed from \OPEN. If $s$ is the goal state, then it is returned. Otherwise, $s$ is expanded (Lines~\ref{expansion:start}--\ref{expansion:end}): for each successor of $s'$ of $s$ that is such that the newly found path through $s$ has lower $g$-value than any previously found path to $s'$, we insert it to \OPEN, and also to \FOCAL if $f(s)\leq wf_{min}$. Since the value of $f_{min}$ may increase during execution, and thus nodes that previously were added to \OPEN but not to \FOCAL, may have to be moved to \FOCAL (Line~\ref{toopen}). This is accomplished by procedure \textit{updateLowerBound}.

\begin{algorithm}[!t]
\small
\DontPrintSemicolon
\SetKwFunction{algo}{algo}\SetKwFunction{proc}{proc}
\KwIn{A search task $P=(G,s_{start},s_{goal})$, an admissible heuristic $h$, a suboptimality bound $w$, a function $h_{\FOCAL}$}
\KwOut{A goal node reachable from $n_{start}$}
\ForEach{$s\in S$}{
    $g(n)\gets \infty$\;
}
$g(s_{start})\gets 0$ \;
$parent(s_{start})\gets \text{null}$\;
$f(s_{start})\gets h(s_{start})$ \;
Insert $n_{start}$ to \OPEN and \FOCAL\;
\While{\FOCAL is not empty}{
    $f_{min} \gets \text{$f$-value of node at the top of \OPEN}$\;
    Extract $s$ from \FOCAL which \emph{maximizes} $h_{\FOCAL}$ \;
    Remove $s$ from \OPEN \;
    \If{$n=n_{goal}$}{
        \Return{$n$}
    }
    \ForEach{$t \in Succ(s)$}{ \label{expansion:start}
        $cost_t \gets g(s) + c(s,t)$\;
        \If{$cost_t < g(t)$}{
        $parent(t)\gets n$ \;
        $g(t) \gets cost_t$\;
        $f(t) \gets g(t)+h(t)$\;
        Insert $t$ into \OPEN\;
        \If{$f(t) \leq wf_{min}$}{
            Insert $t$ into \FOCAL\;
        }
    }\label{expansion:end}
    }
    $top \gets \text{state at the top of \OPEN}$\; 
    \If{$f_{min} < f(top)$}{ \label{toopen}
        \textit{updateLowerBound}$(wf_{min}, wf(top))$
    }
}
\Return{``no solution found''}\;
\;
\SetKwProg{myproc}{procedure}{}{}
    \myproc{updateLowerBound$(old\_{bound}, new\_{bound})$}{
        \ForEach{$s \in \OPEN$}{
            \If{$old\_{bound} < f(n) \leq new\_{bound}$}{
                Insert $n$ into \FOCAL\;
            }
        }
    }
\caption{{\sc Focal Search}}
\label{algo:fs}
\end{algorithm}

\section{Policies and Synthetic Policies}
Recall that our motivation is that current ML techniques allow training policies for search problems.  In this section we specify the requirements that such learned policies should satisfy. Furthermore, since our objective is to evaluate our heuristics for as many accuracy configurations as possible, we describe a simple approach to creating synthetic policies with a given target accuracy.

\subsection{Requirements for Learned Policies}

We assume our policies are constructed using a neural network with an output layer of dimension $|A|$. Intuitively, given a search state $s$, the network returns number between 0 and 1, for every action in $A$. We do not impose any restrictions on the way we encode a search state into an input vector but we do assume that the features of the last hidden layer is given by an $|A|$-dimensional vector $\bm{h}$, defined by:
\begin{equation}
    \label{eqn:logit}
    \bm{h}=f(\bm{x}^s;\bm{\theta}),
\end{equation}
where $\bm{x}^s$ is an input vector representation of a search state $s$,  $\bm{\theta}$ are the parameters of the net, and $f$ encodes the way in which $\bm{\theta}$ operates over $\bm{x}$. Finally, the output layer is an $|A|$-dimensional vector $\bm{y}$, whose components are defined in terms of the \emph{softmax function}, as follows:
\begin{equation}
    \label{eqn:softmax}
    y_i = \frac{\exp(h_i)}{\sum_{j=1}^{|A|}\exp(h_j)}.   
\end{equation}
Given that softmax is that used for the output layer, the sum of the components of $\bm{y}$ is 1, and thus may be interpreted as probabilities \cite{Bridle89}. Now we assume that actions in $A$ can be indexed, by associating each action with a unique identifier between $1$ and $|A|$, using function $\mathit{index}:A\rightarrow \{1,\ldots,|A|\}$. From the network, we can define a stochastic policy as follows:
\begin{equation}
    \pi(a,s)=y_{\mathit{index}(a)}
\end{equation}
We do not require that $\pi$ be well-defined.

\subsection{Generating Synthetic Policies}
To train a policy satisfying the requirements above, one needs to define an adequate representation of the input and an architecture. Then, we choose a method to train the policy. For example, if we use imitation learning \cite{RossGB11}, we generate a number of examples containing elements of the form $(s,a)$, where $s$ is a search state and $a$ is an action in $opt(s)$. After training the policy---a time-consuming process---we can estimate the policy's  accuracy using the test set. Such an accuracy, however, depends on a number of factors, including the representation for the input, the architecture, and the number of examples.

Since in our empirical evaluation we aim at evaluating the heuristics, but not the learning approach, we use a simple method to generate \emph{synthetic stochastic policies}. A synthetic stochastic policy is a function $\pi:A,S\rightarrow [0,1]$ satisfying all the constraints of a stochastic policy, but that is not obtained by training. An advantage of this approach is that it receives a \emph{target accuracy} as input. A disadvantage is that it can only be implemented for search tasks whose search space fits in memory.

Given a target accuracy $acc$, we generate a synthetic policy in two steps. In the first step, we build a table $OPT$ with one pair $(a_{opt},s)$ for every state $s$ in the search space $S$, where $a\in opt(s)$. To do this we run Dijkstra's algorithm from the goal state. Using the softmax function applied over a set of $|A|$ random numbers, we generate a list $y_1,y_2,\ldots,y_{|A|}$ of scores sorted in descending order. In the second step, for each pair $(a_{opt},s)$ in $OPT$, we define $\pi(a_{opt},s)=y_1$ with probability $acc$; and we assign $\pi(a_{opt},s)=y_j$, for some $j\geq 2$ with probability $(1-acc) y_j/\sum_{k=2}^{|A|} y_k$. For the rest of the actions in $A\setminus\{a_{opt}\}$, we assign one of the unused scores randomly. Note that this way of generating the policy guarantees that the optimal action gets the highest score with probability $acc$. But it also is such that when the optimal action does not get the highest score, then it is most likely that it will get the second-largest score. In general the probability that it gets the $j$-th largest score decreases as $j$ increases. This is consistent with what in practice happens with a learned policy, in which one would expect that if a mistake is made, then still the correct action gets assigned a score that is higher relative to other (wrong) actions.

\section{Policies and Focal Search}
In this section we present a number of ways a stochastic policy $\pi$ can be integrated into FS. As we mentioned above, some of our heuristics have been used previously in search literature. Thus, the value of this section lies mainly on the mathematical interpretation of the resulting algorithms. For the two types of heuristics we propose---score- and discrepancy-based---we start off defining in what order we would like our FS's expansions to be carried out, and then derive the heuristic that materializes such an order. 

\subsection{Score-Based $h_{\FOCAL}$}
Assume state $s$ has just been generated and that it is about to be inserted into \FOCAL. Assume further that the path---given by the $parent$ relation---from $s_{start}$ to $s$ is $\sigma(s) = s_1,s_2,\ldots, s_n$, with $s_1=s_{start}$ and $s_n=s$. The likelihood that $\sigma(s)$ is generated by unrolling $\pi$ from the initial state is given by the following expression: 
\begin{equation}
    \mathcal{L}_\pi(\sigma(s)) = \prod_{i=1}^{n-1} \pi(\lambda(s_{i},s_{i+1}),s_i)
\end{equation}
Our first way of exploiting $\pi$ is using the following heuristic as $h_{\FOCAL}$:
\begin{equation}
  h^{\text{path}}_{\mathcal{L}}(s)=-\mathcal{L}_\pi(\sigma(s)), \tag{Score-1} 
\end{equation}
where the negative sign is used because \FOCAL is sorted in ascending order.
Using $h^{\text{path}}_{\mathcal{L}}(s)$ for \FOCAL results in an algorithm that expands first those nodes maximizing the likelihood that path $\sigma(s)$ is generated by 
$\pi$.

A potential caveat of $h^{\text{path}}_{\mathcal{L}}$ is that longer paths are at disadvantage with respect to shorter paths, as the likelihood tends to decrease with path length. To account for this, we define the following score-based heuristic:
\begin{equation}
  h^{\text{path}}_{\mathcal{L}/f}(s)= \frac{h^{\text{path}}_{\mathcal{L}}(s)}{f(s)}. \tag{Score-2} 
\end{equation}
Observe here that by dividing by $f(s)$, we make the heuristic larger, and thus $s$ more preferred, when  $f(s)$ is small (and vice-versa when $f(s)$ is large). As such $h^{\text{path}}_{\mathcal{L}/f}(s)$ incorporates a preference for nodes with low $f$-value. This kind of normalization had also been considered in beam-search approaches \cite[e.g.,][]{wu2016googles} with score-based heuristics.

An alternative to evaluate a state that has just been generated from its parent is by simply focusing on the last edge of the path $\sigma(s)$, namely  $(s_{n-1},s_n)$, and consider the likelihood that the stochastic policy  generates $s_n$ from its parent, $s_{n-1}$. With this in mind, we propose the following two heuristics:
\begin{align}
    h^{\text{last}}_{\mathcal{L}}(s) &=  \pi(\lambda(s_{n-1},s_{n}),s_{n-1}),\tag{Score-3}\\
     h^{\text{last}}_{\mathcal{L}/f}(s) &= \frac{h^{\text{last}}_{\mathcal{L}}(s)}{f(s)}. \tag{Score-4} 
\end{align}
These two heuristics can be viewed as ignoring the ``influence of the past'' over $s$, except for the action that immediately produced $s$.


    
\subsection{Discrepancy-Based $h_{\FOCAL}$}
\newcommand{\pprefix}{\ensuremath{p_{\text{prefix}}}\xspace}
\newcommand{\hprefix}{\ensuremath{h_{\text{prefix}}}\xspace}
Assume $s$ has been generated via path $\sigma(s) = s_1,s_2,\ldots s_n$, with $s_0=s_{start}$ and $s_n=s$, and that it is about to be inserted into \FOCAL. Now, consider the probability that $\sigma(s)$ is the prefix of an optimal path, and let us denote this probability by $\pprefix(s)$. If we could use \pprefix to  ``sort'' \FOCAL, we would obtain an algorithm that prefers paths are most likely to be advancing optimally towards the goal state.

To compute \pprefix, let us assume that $a_1\ldots a_{n-1}$ are the actions associated with path $\sigma(s)$ (i.e.,  $\lambda(s_i,s_{i+1})=a_i$, for every $i\in\{0,\ldots, n-1\}$). The value of $\pprefix(s)$ is given by the product $p_1\cdot p_2\cdots p_n$, where $p_i$ is the probability that $a_i$ initiates a path from $s_i$ to the goal state. Formally, $p_i = P(a_i\in opt(s_i))$, which can be rewritten in this way:
\begin{multline}\small
\hspace{-1ex}P(a_i\in opt(s_i))=\\
    P(a_i \in opt(s_i)\mid \delta(s) \in opt(s_i)) \cdot P(\delta(s) \in opt(s_i)) + \\
        P(a_i \in opt(s_i)\mid \delta(s) \not\in opt(s_i)) \cdot P(\delta(s) \not\in opt(s_i)),\\
\end{multline}
Recall that $\delta(s)$ is the deterministic policy that is obtained from our stochastic policy $\pi$, which returns the action associated with highest probability in $s$. Thus the expression above is conditioning on the fact that the policy may have been right or wrong each time a decision was made. 
For the rest of the calculation, to simplify the resulting expression, we make the assumption that $|opt(s_i)|=1$, for every $i$. 
We have two cases for $p_i$:
\begin{enumerate}
    \item $a_i = \delta(s_i)$; that is $a_i$ is the action returned by the policy at state $s_i$. Then,
    \begin{align}
        P(a_i \in opt(s_i)\mid \delta(s) \in opt(s_i))&=1, \label{eq:p1}  \\ 
        P(a_i \in opt(s_i)\mid \delta(s) \not\in opt(s_i))&=0. \label{eq:p2}
    \end{align}
    Indeed, \eqref{eq:p1} holds since we condition on the fact that $\delta(s_i)$ does return an optimal action in $s_i$. Equation \eqref{eq:p2} holds because we assume the policy does not return an optimal action in $s_i$.
    \item $a_i \neq \delta(s_i)$; that is $a_i$ is not the action returned by the policy at state $s_i$. Then,
    \begin{align}
        P(a_i \in opt(s_i)\mid \delta(s) \in opt(s_i))&=0, \label{eq:p12}\\
        P(a_i \in opt(s_i)\mid \delta(s) \not\in opt(s_i))&=1/\alpha \label{eq:p22}
    \end{align}
    Equation \eqref{eq:p12} holds because of our assumption that $|opt(s_i)|=1$, and that we condition on the fact that the policy does return the (only) optimal action in $s_i$. For Equation \eqref{eq:p22}, observe that we condition on the fact that the policy does not return an optimal action and assume that $a_i$ is \emph{not} returned by the policy. We assume, then, that all remaining actions have the same probability of being optimal and define $\alpha=|A|-1$
\end{enumerate}
Summing up, $p_i$ is as follows:
\begin{equation}
    p_i=\begin{cases} 
            acc_{\pi} & \text{if $a_i=\delta(s_i),$}\\
            \frac{1-acc_{\pi}}{\alpha} &\text{otherwise}.
        \end{cases}
\end{equation}
Now we define $N_{\text{pref}}(\sigma(s))$ as the number of times along the path $\sigma(s)$ in which the action preferred by $\delta$ was taken (i.e., $N_{\text{pref}}(\sigma(s))=\sum_{i=1}^{n-1} [a_i=\delta(s_i)]$), and $N_{\text{nonpref}}(s)$ is the number of times an action not preferred by $\delta$ was taken.
\begin{equation}\label{eq:pprefix}
    \pprefix(s) = acc_{\pi}^{N_{\text{pref}}(\sigma(s))}(\frac{1-acc_{\pi}}{\alpha})^{N_{\text{nonpref}}(\sigma(s))},
\end{equation}
 Now we define $h_{\text{prefix}}$ by taking the logarithm of $\pprefix$ and dividing by $\log(\frac{1-acc_{\pi}}{\alpha})$:
\begin{equation}\label{eq:hprefix}
    h_{\text{prefix}}(s) = \frac{\log(acc_{\pi})}{\log(\frac{1-acc_{\pi}}{\alpha})} N_{\text{pref}}(\sigma(s))  + N_{\text{nonpref}}(\sigma(s))  \tag{Disc-1}
\end{equation}
Note that even though we take the logarithm of a value between 0 and 1, by dividing by the constant negative number $\log(1-acc_{\pi})$ we restore the original order; that is, $h_{\text{prefix}}$ grows exactly when $\pprefix$ grows.

\subsubsection{From Probability to Discrepancy} Note that the ratio $\frac{\log(acc_{\pi})}{\log(\frac{1-acc_{\pi}}{\alpha})}$ in \eqref{eq:hprefix} approaches 0 as the accuracy of the policy $acc_{\pi}$ approaches 1.
For example, if our problem has branching factor 4, and policy $\pi$ has an accuracy of 90\%, the ratio equals to 0.02. Furthermore the ratio also decreases as the branching factor increases. If we assume that our policy is very accurate or, alternatively, that the branching factor is high, we can remove the first term of $h_{\text{prefix}}$ to produce a much simpler expression:
\begin{equation}\label{eq:hdisc}
    h_{\text{disc}}(s) = N_{\text{nonpref}}(\sigma(s)). \tag{Disc-2}
\end{equation}
An algorithm that uses $h_{\text{disc}}$ to sort \FOCAL, therefore, is one that prefers for expansion a state $s$ that maximizes an approximation, which assumes $\pi$ is accurate, of the probability that the path to $s$ is a prefix of an optimal solution.

$h_{\text{disc}}(s)$ is an analogue of the notion of \emph{discrepancy} of the Limited Discrepancy Search algorithm \cite{HarveyG95}. As originally conceived by \citeauthor{HarveyG95}, the notion of discrepancy is defined for a path of states $\sigma=s_1,\ldots, s_n$. To compute $\sigma$'s discrepancy we initialize our discrepancy counter to zero, iterate an index $i$ from 2 to $n$, and increment the counter when there is a successor of $s_{i-1}$, different from  $s_i$, which has an $h$-value lower than $h(s_i)$. Instead of using a heuristic function to count discrepancies, $h_{\text{disc}}(s)$ uses the policy $\pi$. 

Discrepancies were originally proposed for binary trees. Some researchers \cite[e.g.,][]{KarouiHLN07} have considered counting discrepancies according to their successor rank in non-binary trees. In our experimental section we evaluate a variant of $h_{\text{disc}}$, defined as follows:
\begin{equation}\label{eq:hdiscrank}
    h_{\text{rank}}(s) = \mathrm{rank}(\sigma(s)) \tag{Disc-3}
\end{equation}
where $\mathrm{rank}(\sigma(s))$ computes the rank of the action used to produce $s$ from its predecessor, $s_{n-1}$, in the stochastic policy $\pi(\cdot,s_{n-1})$. The rank of the first action is 0.

\section{Empirical Evaluation}
Our empirical evaluation had two objectives. The first was to evaluate the performance of the proposed heuristics over different domains, using them along with policies of varying accuracy. The second objective was to see whether or not the conclusions obtained with small domains may apply to larger ones. Hence, we divide this section into two parts. First, we exhaustively evaluate the performance of our heuristics under various accuracy settings, using our approach to compute synthetic policies, over three domains. Second, we evaluate the 15-puzzle with a learned policy.

In the evaluation we include two other algorithms: Weighted A* (WA*) \cite{Pohl70}, and A* with preferred  operators (PrefA*), a variant of Fast-Downward's  search algorithm \cite{helmert06}, where, intuitively, the action selected by the policy is a preferred operator. Specifically, PrefA* uses two open lists: \emph{preferred} open and \emph{regular} open, and  adds the successor of $s$ generated via action $\delta_\pi(s)$ to \emph{preferred}, while all other successors are added to \emph{regular}. For expansion, it prefers to extract a node from  \emph{preferred} always, unless this list is empty (we tried alternation, like in the original algorithm, and found that no alternation led to a faster algorithm). Note that \emph{preferred} may get empty when the policy leads to state revisiting. Both open lists are ranked by $f=g+h$.   PrefA* is not a bounded-suboptimal algorithm; it is included for reference since it allows us to estimate the quality of the solutions that would be obtained by unrolling the policy while still remaining complete.

The experiments with synthetic policies were run on an Intel i5 8250U PC with 8GB RAM. The trained policy experiments were run on an Intel Xeon E5-2630 machine with 128GB RAM, using a single CPU core and no GPU. We use labels \emph{Score-1..4} and \emph{Disc-1..3} to identify our 4 score-based heuristics and 3 discrepancy-based heuristics, respectively.

\subsection{Synthetic Policies}
According to the methodology presented, in which for each domain we select a single goal state, and solve the entire search space, we generate synthetic policies with target accuracy in $\{70\%, 80\%, 90\%, 95\%, 100\%\}$. The domains we use and the results are described below.

\Paragraph{8-Puzzle}
This is the $3\times 3$ version of the classic sliding-tile puzzle \cite[e.g.][]{Korf85}. Our experiments use the Linear Conflicts heuristic \cite{HanssonMY92}.

The first two rows of Figure \ref{fig:small} show the runtime, expansions, and suboptimality (computed as the ratio between the cost of the returned solutions and the optimal cost), using two different suboptimality bounds: $1.2$ and $1.5$. WA* is displayed as a horizontal line because its performance is policy-independent.
Heuristics \emph{Score-3} and \emph{Score-4} seem to be less competitive than other algorithms, while \emph{Disc-2} and \emph{Disc-3}, are superior to other configurations both in the number expansions and solution quality. WA* is competitive when the suboptimality bound is $1.2$. PrefA* is the fastest but returns solutions not that exceed the suboptimality bound.

\Paragraph{Pancake Sorting}
The problem consists of an array that must be sorted. Each action corresponds to flipping the array from a certain position. Our experiments were conducted using 9 pancakes, yielding a total of $9!$ ($362,880$) reachable states. We use the gap heuristic \cite{DBLP:conf/socs/Helmert10}.
The third and fourth rows of Figure~\ref{fig:small} show the results (runtime, expansions, and suboptimality) that were obtained. In this domain, \emph{Score-3} and \emph{Score-4} outperform other algorithms on low accuracy policies, but \emph{Disc-1} and \emph{Disc-2} outperform other algorithms as accuracy improves. Nevertheless, in this domain, we could observe that WA* is faster than other algorithms, even though it perfoms more expansions. This is due to an inherent disadvantage of FS over WA*, which needs to maintain two open lists. PrefA* is the fastest but returns solutions not that exceed the suboptimality bound.

\Paragraph{Blocks World}
This is the 4-operator blocks world from IPC-2000. The problem consists of a set of blocks, a table, and a robotic arm. Each block can be over another block or the table. The robotic arm can hold one block or be empty. The goal consists of finding a plan that transforms one configuration of blocks into another. We used instance number 13 of the blocksworld from IPC (probBLOCKS-8-0.pddl), in which the objective is to build a single 8-block tower. We use $h_{\max}$  \cite{bongeff00} as the heuristic.

The last two rows of Figure~\ref{fig:small} show the results (runtime, expansions, and suboptimality) obtained. We observe that \emph{Disc-2} and \emph{Disc-3} outperform all other algorithms in terms of time and expansions; nevertheless, solutions obtained are just within the bound. \emph{Disc-2} and \emph{Score-2}, yield higher-quality solutions. In this domain, the poor performance of WA* may be due to the fact that the heuristic is not informed compared to other domains. PrefA* performs poorly when the policy has a low accuracy. While the policy increases its accuracy, PrefA* becomes faster, but its solutions exceed the suboptimality bound.

In summary, in all domains we observe that FS, used with \emph{Disc-2} and \emph{Disc-3}, yields results that are superior to other algorithms. We observe that in domains in which the heuristic is more informed (i.e., pancake sorting) WA* is more competitive. We observe that if the policy accuracy is at least 80\%, it is worth using FS with the policy rather than WA*. 

Our conclusions cannot be extended to the case of a very low suboptimality bound (i.e., very close to the limit $w=1$), since in those cases WA* is faster than FS. This is because, to prove suboptimality, both algorithms need to expand a similar number of nodes, and FS incurs in a greater overhead by maintaining two priority queues.

  \newcommand{\ww}{0.97}
\begin{figure*}
    \centering
    
    \includegraphics[width=\ww\textwidth,clip]{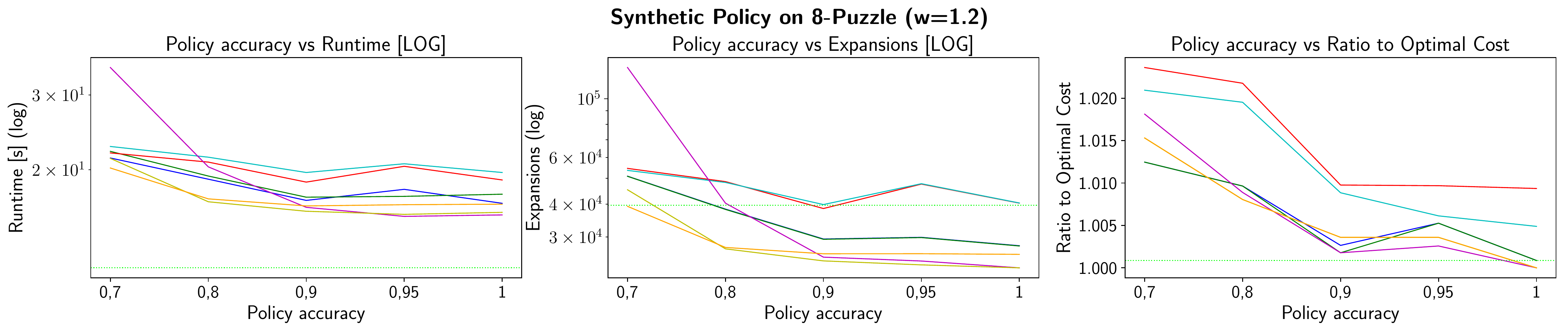}
    \includegraphics[width=\ww\textwidth,clip]{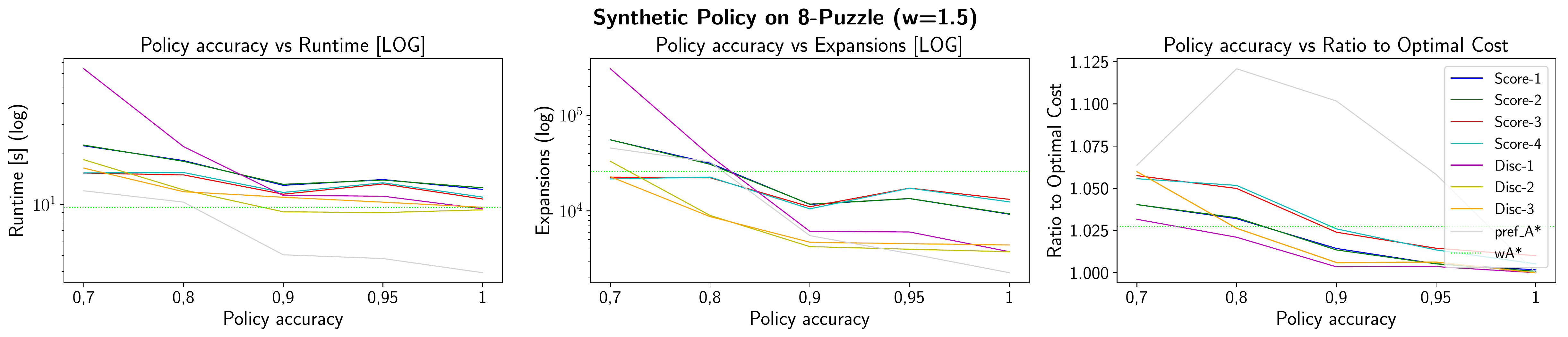}

    \includegraphics[width=\ww\textwidth,clip]{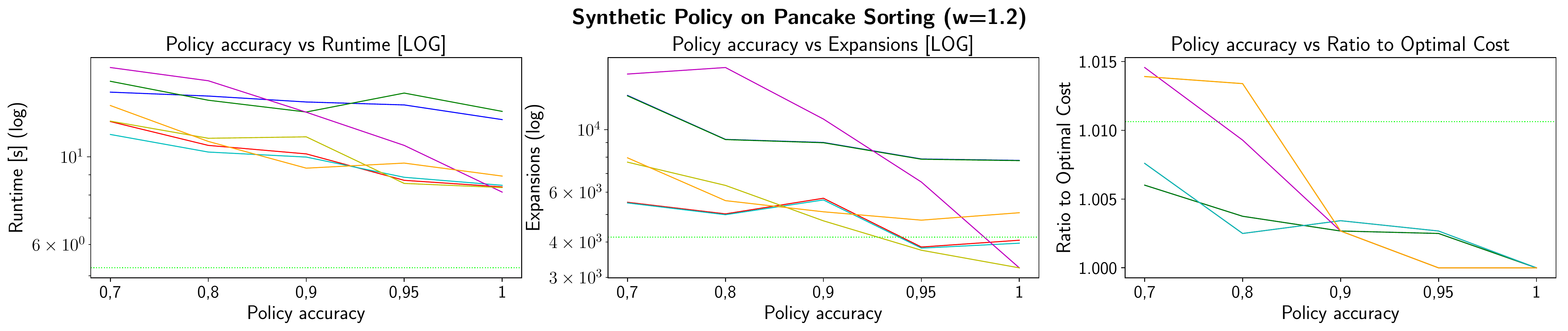}
    \includegraphics[width=\ww\textwidth,clip]{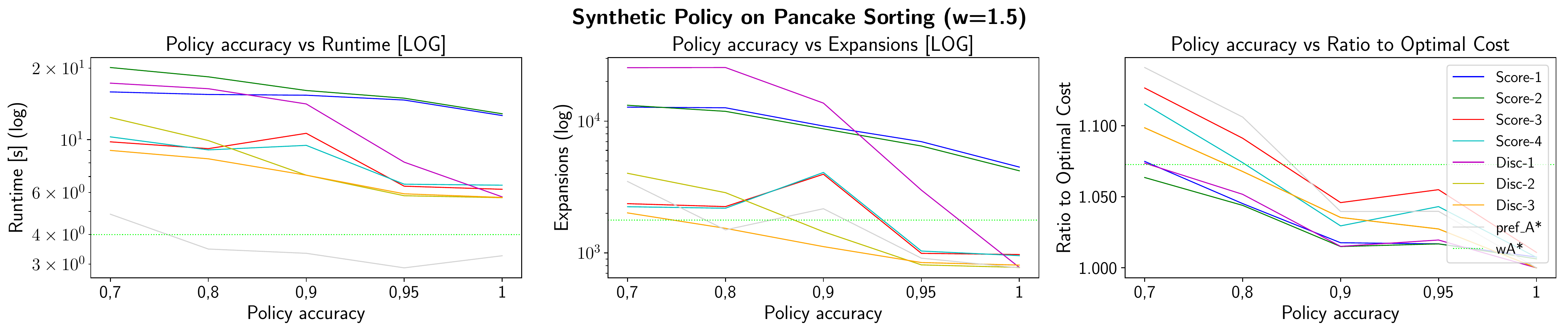}

    \includegraphics[width=\ww\textwidth,clip]{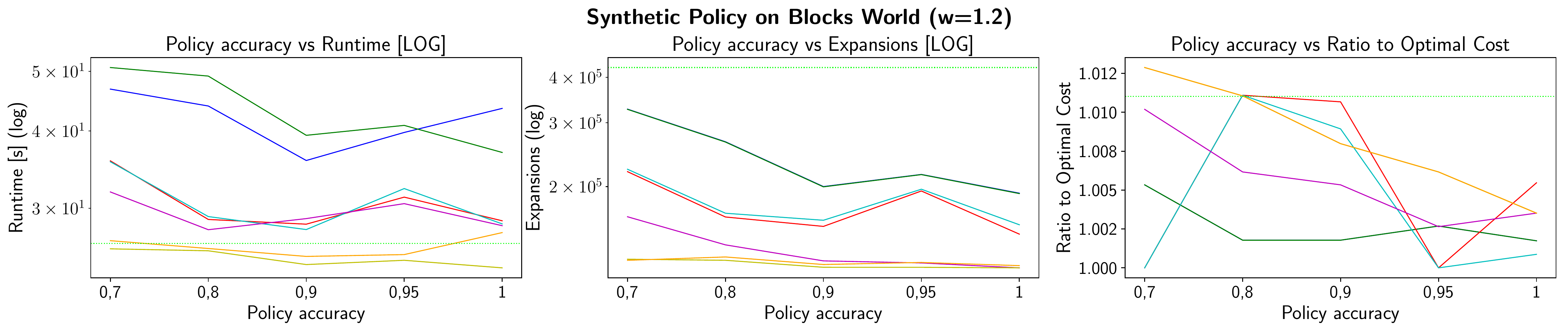}
    \includegraphics[width=\ww\textwidth,clip]{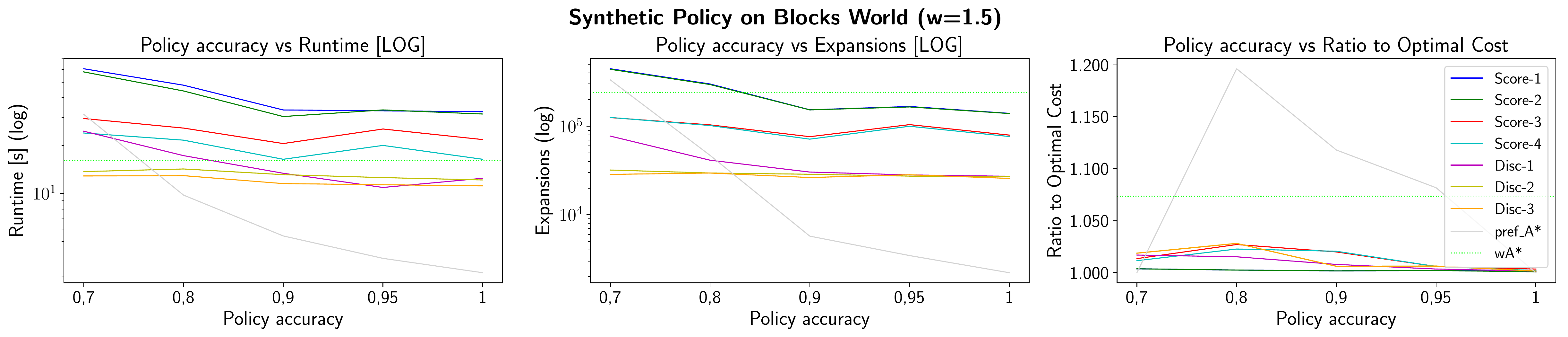}
    \caption{Results for 8-puzzle, pancake sorting, and blocksworld}
    \label{fig:small}
\end{figure*}

\subsection{15-Puzzle with a Trained Policy}
\newcommand{\www}{0.95}
\begin{figure*}[t]
    \centering
    \includegraphics[width=\www\textwidth,clip]{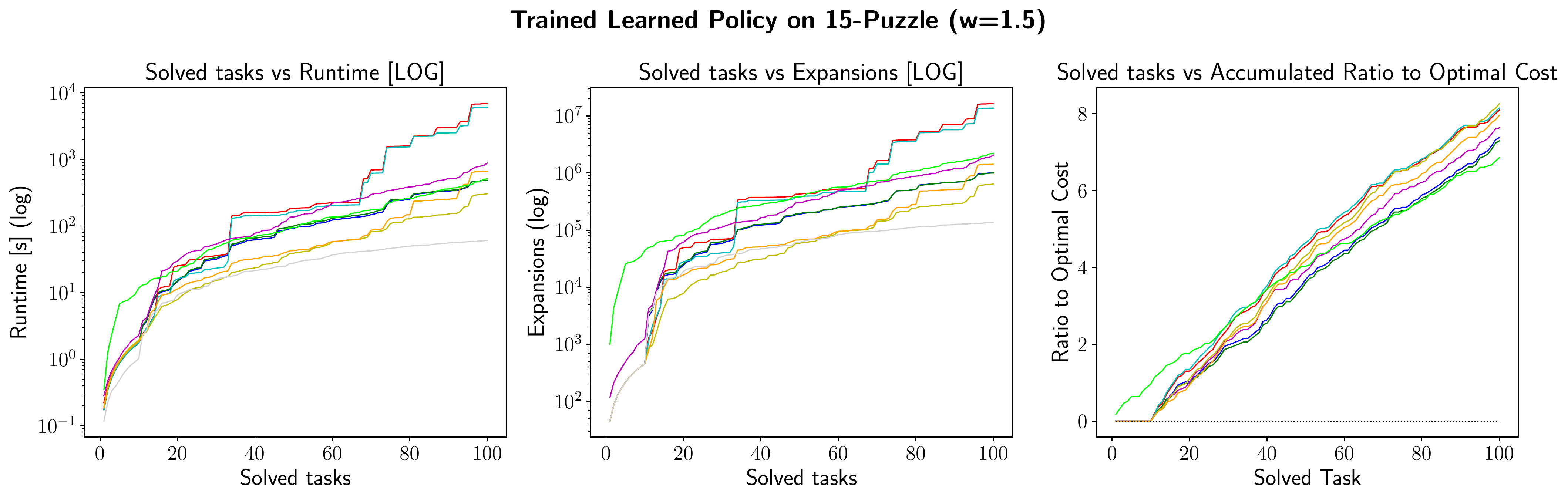}
    \includegraphics[width=\www\textwidth,clip]{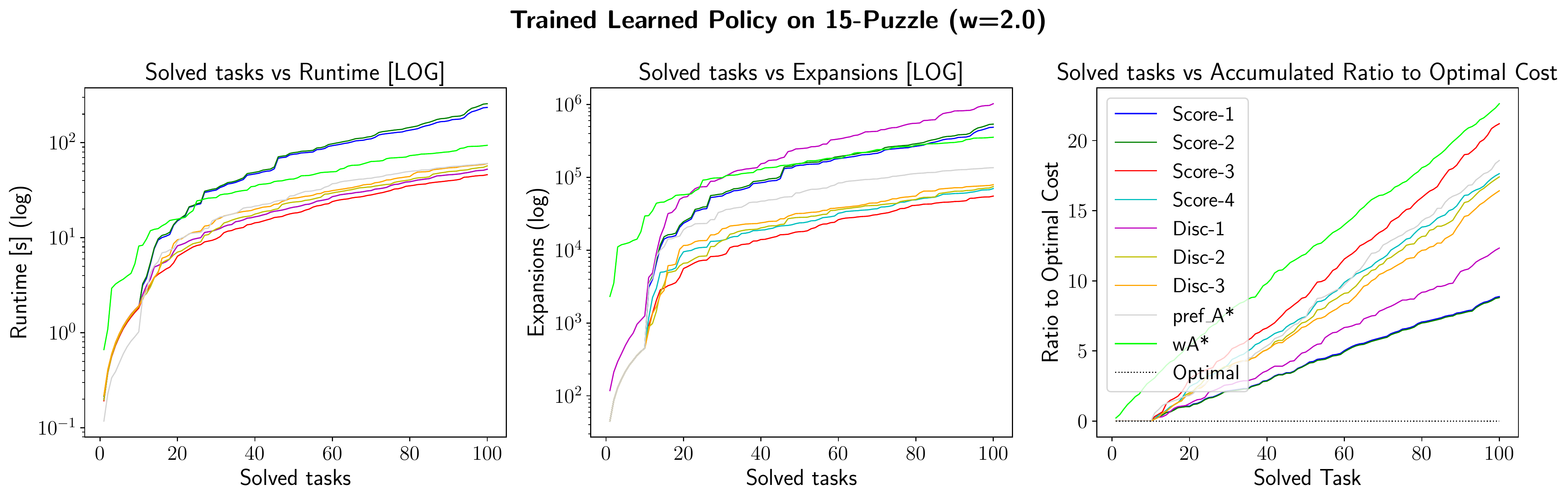}
    \caption{Results using a trained learned policy in 15-puzzle over Korf's 100 instances. First row shows results with a suboptimality bound $w=1.5$; and the second row shows results with suboptimality bound of $w=2.0$.}
    \label{fig:results_learned}
\end{figure*}

We trained a simple neural network that we generated with our own data set of 1,510,673 examples extracted from 30,367 optimal traces, obtained by solving randomly generated problems with IDA* \cite{Korf85}. The dataset was split in 90\%-10\% for the train and test sets, respectively. The architecture of the neural network has an input layer with 256 neurons, where each tile in the puzzle is represented in a one-hot vector; three hidden layers of 160, 80, and 16 neurons, and, finally, a 4-dimension output layer with softmax as activation function, where each one represents an action. 
After training the neural network for 200 epochs, minimizing the cross-entropy loss \cite{goodfellow2016deep}, the policy's accuracy was 87.5\% on the test set.

To evaluate the algorithms' performance using the learned policy, we use  Korf's 100 search tasks \cite{Korf85}. We run our algorithm using the Linear Conflicts heuristic \cite{HanssonMY92}.

Figure \ref{fig:results_learned} shows the results---in terms of runtime, expansions, and accumulated suboptimality---using the learned policy with the suboptimality bound set to $2.0$ and $1.5$. To compute accumulated suboptimality, for each solved task we compute the ratio of the cost of the returned solution and the optimal cost and subtract one; then we add this amount for every solved task given a timeout. For $w=2.0$, the results show that \emph{Score-3}, \emph{Score-4}, \emph{Disc-2}, and \emph{Disc-3} outperform WA* by one order of magnitude with respect to the number of expansions and half an order of magnitude regarding runtime. Regarding the accumulated suboptimality, \emph{Score-1} and \emph{Score-2} obtain solutions closer to the optimal solution, but \emph{Score-4}, \emph{Disc-2}, and  \emph{Disc-3} return better-quality solutions than PrefA* and WA*. 
For $w=2.0$, we observe that \emph{Disc-2}, \emph{Disc-3}, and \emph{Score-2} outperform WA* regarding the number of expansions, although only \emph{Disc-2} outperforms WA* in terms of runtime. This is due to the fact that expansions are slower when using our trained policy because evaluating the neural net requires significant computation. Indeed, on average, an expansion is one order of magnitude slower using the learned policy.



Finally, PrefA* is that it solves very few problems within suboptimality bound $w=1.5$, showing that simply unrolling the policy yields poor quality solutions in general. 

In summary, the results show that \emph{Disc-2} and \emph{Disc-3}, the most straightforward way to include discrepancies in FS, outperform all other algorithms in terms of runtime and expansions, making a good trade-off between expansions and solution quality within the suboptimality bound. Nevertheless, \emph{Score-3} and \emph{Score-4} shows good results when a higher suboptimality bound is required, decreasing its performance when the suboptimality bound decrease.

\section{Discussion}
An important limitation of this work is that an important portion of our evaluation is carried out on small problems whose search space fits in memory. We chose this approach because it allowed us to be exhaustive with respect to the accuracy of the policy being used. Nevertheless, approaches for the same type of analysis over tasks with larger search space do exist. For example, one could choose different neural network architectures, train them over a large number of examples for a given family of tasks. This would yield different estimated accuracies over the test set. By carefully tuning the architecture, policies with different accuracies can be obtained. While this is an interesting line of work, it requires plenty of hand tuning, shifting the focus away from the search algorithms, which was the original objective of our research. Notwithstanding, our results for the 15-puzzle are consistent with those obtained on smaller tasks, suggesting that our conclusions may extend to other larger domains.

Another limitation of our synthetic approach to the generation of policies is that we consider the policy's accuracy to be distributed homogeneously on the state space. However, this may not be the case for  trained learned policies. For example, our policy for the 15-puzzle is able to solve optimally Korf's 10 first problems, but does not produce good-quality results for the harder tasks. Modifying the synthetic approach to include a greater proportion of errors for states that are farther away from the goal is easy to do; in this case, a question to answer is whether or not there exist heuristics capable of effectively exploiting this characteristic.

Two pairs of our score-based heuristics obtain similar results (\emph{Score-1} vs. \emph{Score-2}, and \emph{Score-3} vs. \emph{Score-4}). This is not surprising since \emph{Score-2} is obtained directly from \emph{Score-1}, while the same happens with the other pair. We decided to still include \emph{Score-2} and \emph{Score-4} in our analysis, since in previous research it had been shown that score normalization produced better-quality results \cite{wu2016googles}. While establishing a good tradeoff between scores and solution quality seems important, it does not seem that our specific approach is the correct one. This may be due to the fact that FS already controls suboptimality using the heuristic. Further research is necessary to establish a more effective way to incorporate this tradeoff.



Finally, in our experimental comparison, we used WA* as the main bounded-suboptimal algorithm for comparison, instead of comparing to more recent bounded-suboptimal algorithms like Explicit Estimation Search (EES) \cite{ThayerR11}. We do this since our initial focus was on domains with unitary cost, which are most abundant in the search literature. EES does not significantly outperform WA* in these types of domains \cite{ThayerR11}. A study considering non-unitary costs is an interesting line of future research. Integration and comparison with Improved Optimistic Search \cite{ChenSDR19}, an approach to suboptimal search also based on Focal Search, is also an interesting line of future research.




\section{Conclusions}
We presented two families of heuristics applicable to Focal Search when a (learned) policy is available. Our mathematical interpretation of these families concluded that score-based heuristics yield a search algorithm that expands a node which, among all other nodes in \FOCAL, is the most likely to have been generated by unrolling the policy. Discrepancy-based heuristics, instead, yield an algorithm that expands a node which, among all other nodes in \FOCAL, maximizes the probability that its path is a prefix of an optimal path. 

We evaluated the heuristics over four domains. In three domains (8-puzzle, blocksworld, pancake) we control the accuracy of the policy and report the performance of each of our heuristics. In the fourth domain (15-puzzle) we compare our heuristics using a trained 87.5\%-accurate policy trained over 1.5 million examples. Consistently, we observe that discrepancy-based heuristics, yield the best results in terms of expansions and runtime, outperforming Weighted A* when the weight used is 1.2 or greater. Score-based heuristics are not fast, but yield the best-quality solutions.

The effectiveness of our approach depends on both the accuracy of the policy and the accuracy of the heuristic. As the accuracy of the policy increases, we observed an increased benefit of using FS; specifically, in our controlled experiments, it is always worth to use FS when the policy's accuracy is 80\% or larger. Exploiting policies yields more benefits when the heuristics are less informed. Finally, when the suboptimality bound approaches 1, the effectiveness of FS relative to WA* decreases, since the number of expansions required by both algorithms becomes similar, and FS requires maintaining two open lists.

\section{Acknowledgements}
We would like to thank Martin Alamos who participated in the initial part of this research.
Matias Greco was supported by the National Agency for Research and Development (ANID) / Doctorado Nacional / 2019 - 21192036.

\bibliography{refs}

\end{document}